%% file: main.tex
\documentclass[10pt,twocolumn,letterpaper]{article}

\usepackage{cvpr}      
\usepackage{graphicx}
\usepackage{booktabs}

\usepackage{lipsum}
\usepackage{multirow}
\usepackage{colortbl}
\usepackage{amsmath}
\usepackage{caption}
\usepackage{verbatim}  
\usepackage{caption}
\usepackage{makecell}
\usepackage{pifont}
\newcommand{\cmark}{\ding{51}} 
\newcommand{\xmark}{\ding{55}} 
\usepackage{tcolorbox}
\tcbuselibrary{breakable}
\newtcolorbox{promptbox}[1][]{
  breakable,
  colback=blue!3,
  colframe=blue!70,
  boxrule=0.4pt,
  arc=2pt,
  left=4pt,
  right=4pt,
  top=4pt,
  bottom=4pt,
  fontupper=\ttfamily\small,
  title=#1
}

\input{preamble}
\definecolor{cvprblue}{rgb}{0.21,0.49,0.74}
\usepackage[pagebackref,breaklinks,colorlinks,allcolors=cvprblue]{hyperref}

\def\paperID{*****} 
\def\confName{CVPR}
\def\confYear{2026}

\title{MegaStyle++: Scaling Image Style Space through Hierarchical Style Definition}

\author{Junyao Gao$^{1,2}$\textsuperscript{*} \quad Sibo Liu$^{2}$\textsuperscript{*} \quad Jiaxing Li$^{3}$ \quad Yanan Sun$^{4}$\\  Weidong Zhang$^{2}$ \quad Jun Zhang$^{2\ddag}$ \quad Cairong Zhao$^{1\ddag}$ \\
	$^{1}$Tongji University, $^{2}$Tencent, $^{3}$Nanyang Technological University, \\ $^{4}$Hong Kong University of Science and Technology\\
	}

\begin{document}
\maketitle

\begingroup
\renewcommand\thefootnote{}
\footnote{Work done during Junyao Gao's internship at AIPD, Tencent. \textsuperscript{\ddag}Corresponding authors. *Equal contributions.}
\addtocounter{footnote}{-1}
\endgroup

\input{sec/0_abstract}    
\input{sec/1_intro}
\input{sec/2_what_is_style}
\input{sec/3_method}
\input{sec/4_experiments}
\input{sec/5_conclusion}
{
    \small
    \bibliographystyle{ieeenat_fullname}
    \bibliography{main}
}


\end{document}

%% file: sec/0_abstract.tex
\begin{abstract}
Image style is a highly abstract, human-constructed concept shaped by a range of visual factors and intrinsically entangled with content, yet a unified and explicit definition of image style remains lacking.
In this work, we first discuss the fundamental question of what is style and then propose a hierarchical style definition that describes image style from an overall style identity to fine-grained visual attributes, providing a more structured, transferable, and interpretable style representation.
Based on this definition, we refine the style annotation pipeline of MegaStyle and construct MegaStyle++-8M, a large-scale style dataset containing 150K overall style identities, 1M fine-grained style prompts, and 8M stylized images.
Extensive analyses demonstrate that our hierarchical definition substantially expands the style space in both diversity and semantic breadth, while precisely capturing intrinsic visual style of reference images.
The dataset and code will be updated at \url{https://github.com/Tencent/MegaStyle}, we hope MegaStyle++ provides a scalable foundation for studying and modeling diverse image styles.
\end{abstract}

%% file: sec/1_intro.tex
\section{Introduction}
\label{sec:intro}
Image style transfer aims to transfer the visual style of a reference image to user-specified content.
It has been widely adopted in everyday applications, such as social media filters and digital art creation.
Despite its rapid development and impressive performance, the core concept underlying this task remains ambiguous \cite{somepalli2024measuring}:

\textit{What is style?}

Many prior works define image style through visual representations extracted from reference images \cite{gatys2016image, ulyanov2016texture,yang2023zero,wang2023styleadapter,ahn2024dreamstyler,qi2024deadiff}.
Specifically, \cite{dumoulin2016learned, johnson2016perceptual} and \cite{11165480,liu2023stylecrafter,xing2024csgo} represent style as image features from pre-trained vision encoders, such as CNNs and CLIP.
Another line of work \cite{everaert2023Diffusion, lu2023specialist, gal2022image, zhang2023inversion} encodes style into model-specific parameters like embeddings \cite{zhang2023inversion,gal2022image}, adapters \cite{sohn2024styledrop} and LoRA weights \cite{hu2021lora,ruiz2023Dreambooth}.
These implicit style definitions are black-box and often capture reference-specific biases beyond style, such as semantic content and encoder priors, leading to entangled style representations.

On the other hand, people rely on natural language to describe visual style, but these explicit definitions are highly subjective and lack a unified standard.
For example, ``watercolor style'' may refer to a painting medium and its associated visual effects, ``Monet style'' may indicate artist-specific brushwork and color usage, and ``abstract style'' may denote a high-level visual abstraction.
In practice, natural-language style descriptions often rely on short and coarse terms, failing to provide precise and distinctive image style descriptions.
Images labeled with the same style can vary significantly in color, texture, and brushwork.
Therefore, the style transfer community urgently needs an explicit and fine-grained style definition to identify the transferable style factors in a reference image and establish clearer and more effective guidance for style transfer.

In this paper, we address this need by introducing a unified style definition that uses hierarchical language instructions to describe image style from an overall style identity to fine-grained visual attributes.
Specifically, we first initialize the style definition of a reference image from its holistic visual impression, and then refine it through detailed style attributes, including color, lighting, texture, artistic medium and brushwork.
This design is grounded in the inherent nature of style: it is an abstract visual concept that emerges from complex interactions among multi-level visual attributes.
Based on this definition, we refine the style instruction prompts in MegaStyle \cite{gao2026megastyle} to scale the image style space, and follow its data curation pipeline to build MegaStyle++-8M, a large-scale style dataset containing 1M fine-grained styles and 150K overall artistic style identities.

Comprehensive analyses demonstrate that our method substantially expands the image style space in both breadth and granularity, enabling more diverse and distinctive style supervision compared with MegaStyle.
More importantly, our hierarchical style definition provides an explicit language for describing style, making the ambiguous notion of style more structured and interpretable, and offering practical value to the style transfer community.
The contributions of this paper are summarized as follows:
\begin{itemize}
    \item We propose the first unified and hierarchical style definition that uses explicit language instructions to describe image style from an overall style identity to fine-grained visual attributes.  
    \item Based on the proposed definition, we scale the image style space in both breadth and granularity, and construct MegaStyle++-8M with 1M fine-grained style prompts organized under 150K overall artistic style identities.
\end{itemize}

%% file: sec/2_what_is_style.tex
\section{What is Style?}
As discussed in \cite{somepalli2024measuring}, the precise definition of image style remains in contention, and no universally accepted definition has yet emerged.
This is because style is a highly abstract, human-constructed concept shaped by a range of visual factors and intrinsically entangled with content.
People often describe style only from a particular perspective, depending on the requirements of different tasks.

Early studies \cite{gatys2016image, ulyanov2016texture, dumoulin2016learned, johnson2016perceptual} represented style using spatially invariant statistics of CNN features.
For example, \cite{gatys2016image} used Gram matrices to capture inter-channel correlations, while AdaIN \cite{huang2017arbitrary} and WCT \cite{li2017universal} matched channel-wise means and variances, and full covariance matrices, respectively.
However, these statistical formulations tend to identify style with texture-like appearance, failing to capture its broader visual complexity.
More recently, \cite{ahn2024dreamstyler,ye2023ip,xing2024csgo,liu2023stylecrafter} leverage more expressive visual representations from large vision-language models (VLMs) like CLIP \cite{radford2021learning} and SigLIP \cite{zhai2023sigmoid} to learn style.
Yet, style and semantic content are highly coupled in VLMs' feature space, making it very difficult to extract clean and rich style representations \cite{11165480}.
With the rapid advancement of diffusion models \cite{xia2024diffi2i,xia2024diffusion,croitoru2023diffusion,sun2024create}, many methods \cite{everaert2023Diffusion, lu2023specialist, gal2022image, zhang2023inversion} encode the entire visual style of given images into learnable model parameters, such as textual embeddings \cite{zhang2023inversion}, adapters \cite{sohn2024styledrop}, and LoRA \cite{hu2021lora} weights.
Despite their impressive performance, these methods essentially learn a mixture of style attributes without establishing explicit and interpretable visual semantics.
Moreover, they often encode content-specific information and are tied to the base model on which they are trained, preventing them from serving as a general style definition for the style transfer community.

In explicit language-based formulations, visual styles are most commonly named after genres, artists, or artistic media.
Under such coarse style definitions, images in the same style still exhibit substantial fine-grained style differences.
Recent works \cite{wu2024fiva,ruta2022stylebabel} move beyond single labels and describe style in terms of various visual attributes such as texture, color and other artistic elements rather than single style label.
However, attribute-only descriptions are insufficient to fully characterize a style, as the same attributes may appear across different styles.
For example, the dusty teal color and visible fibrous paper grain may be found in both watercolor illustrations and vintage screen prints, but take on different stylistic meanings in each context.

Therefore, we define image style as a hierarchy comprising an overall visual identity and its fine-grained visual attributes.
Starting from a holistic style anchor, this definition refines the style through concrete visual attributes, yielding a distinctive and precise description of the reference image's style.

%% file: sec/3_method.tex
\begin{figure}
    \centering
    \includegraphics[width=\linewidth]{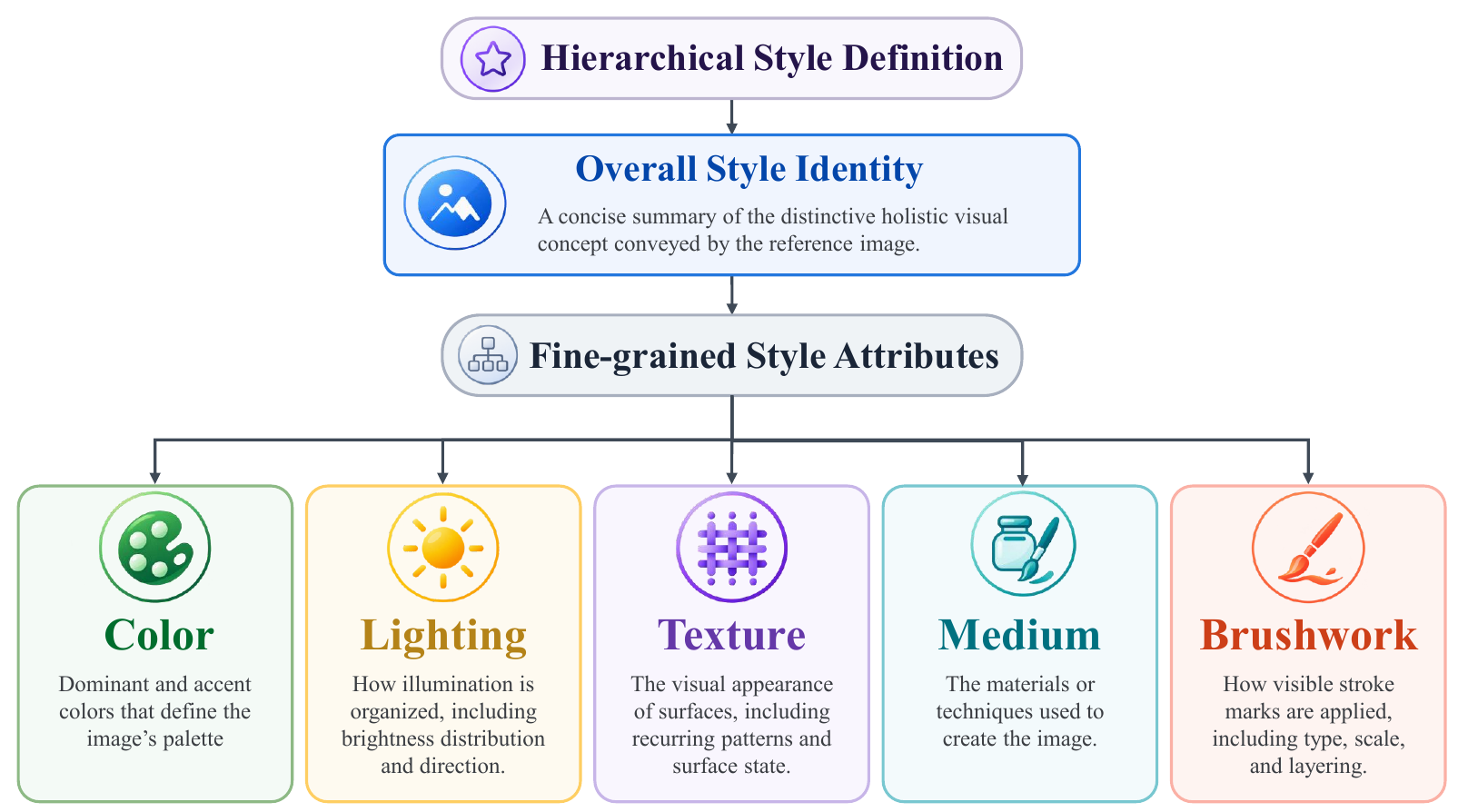}
    \caption{Overview of the proposed hierarchical style definition. Image style is first characterized by an overall style identity that summarizes the distinctive holistic visual concept, and is further refined through five fine-grained style attributes: color, lighting, texture, medium, and brushwork.}
    \label{fig:framework}
\end{figure}

\section{Method}
\label{sec:method}
As illustrated in Figure \ref{fig:framework}, our hierarchical style definition first establishes the overall identity from the holistic visual impression of a reference image and then refines it through style attributes, including color, lighting, texture, medium, and brushwork. 
These attributes capture complementary aspects of style, ranging from global appearance and material qualities to local rendering details.
Moreover, the identified properties should be visually observable, perceptually dominant, and transferable across semantic contents.
To enable large-scale automatic annotation with VLMs, we also design a set of instruction prompts following the proposed style definition.
In what follows, we detail how each style component is defined and identified, together with its corresponding instruction prompt.

\noindent{\textbf{Overall Style Identity.}}
The overall style identity treats image style as a coherent visual concept and provides a concise summary of the holistic impression conveyed by the reference image.
It is not based on fixed objective rules, but arises from the visual experience and conventions accumulated through long-standing artistic practice and perception, and may be explained in terms of genres (e.g., Impressionism) and artists (e.g., Monet).
To determine the overall style identity, we examine the reference image from a global perspective and identify its most distinctive and meaningful holistic visual concept.
And we should avoid listing fine-grained attributes or using generic terms such as ``digital art'' or ``painting''.

\begin{tcolorbox}[
colback=gray!5,
colframe=gray!50,
boxrule=0.5pt,
arc=2pt,
left=6pt,
right=6pt,
top=6pt,
bottom=6pt,
breakable,
title=\textbf{Instruction Prompt: Overall Style Identity},
fonttitle=\small
]
\small\ttfamily
You are a professional image annotator. Please characterize the input image's fine-grained, visually distinctive (not a generic category) overall artistic style based on its intrinsic stylistic attributes, using a single phrase of fewer than four words.
\end{tcolorbox}

\noindent{\textbf{Color.}}
Color is one of the most direct and fundamental style attributes of image style and can be characterized from four aspects: color palette, color properties, color combination, and color distribution.
First, the color palette specifies the dominant and accent colors in the reference image.
Color properties further describe the saturation and temperature of the identified colors.
Color combination describes how neighboring colors perceptually interact, including mutual contrast, assimilation, optical mixing, blending, and overlap.
Color distribution describes the relative proportions and spatial arrangement of the identified colors across the image.
However, some color aspects are primarily determined by other visual factors.
For example, color properties depend strongly on the brightness and spatial distribution of lighting, while color combination and distribution are largely governed by the semantic content and overall style identity of the reference image.
We therefore do not include these aspects in our explicit color annotation.

\begin{tcolorbox}[
colback=gray!5,
colframe=gray!50,
boxrule=0.5pt,
arc=2pt,
left=6pt,
right=6pt,
breakable,
top=6pt,
bottom=6pt,
title=\textbf{Instruction Prompt: Color},
fonttitle=\small
]
\small\ttfamily
You are a professional image annotator. Please identify the dominant colors and accent colors of the input image in four words. When listing colors, output color terms only and do not include any content-related words or descriptions.
\end{tcolorbox}

\noindent{\textbf{Lighting.}}
Lighting describes how illumination is organized across the reference image.
It can be characterized in terms of brightness distribution and direction.
Brightness distribution captures the spatial variation in illumination strength across the image, while lighting direction indicates where the illumination comes from.

\begin{tcolorbox}[
colback=gray!5,
colframe=gray!50,
boxrule=0.5pt,
arc=2pt,
left=6pt,
breakable,
right=6pt,
top=6pt,
bottom=6pt,
title=\textbf{Instruction Prompt: Lighting},
fonttitle=\small
]
\small\ttfamily
You are a professional image annotator. Please characterize the style attribute 'lighting' of the input image in eight words based on the following instructions:
\begin{enumerate}
    \item Identify the primary lighting direction.
    \item Describe the brightness distribution: how its brightness transitions across the image.
    \item Do not include the color of light and any content-related words or descriptions.
\end{enumerate}
\end{tcolorbox}
\noindent{\textbf{Texture.}}
Texture describes the visual appearance of surfaces in an image.
It can be characterized in terms of recurring visual patterns and surface state.
Recurring visual patterns refer to identifiable local structures that repeatedly occur across a surface, together with the scale and form of these structures.
Surface state describes the visible finish and condition of a surface, including effects produced by surface treatments, fabrication processes, and wear, as well as how the surface reflects and scatters light.

\begin{tcolorbox}[
colback=gray!5,
colframe=gray!50,
boxrule=0.5pt,
arc=2pt,
left=6pt,
right=6pt,
top=6pt,
breakable,
bottom=6pt,
title=\textbf{Instruction Prompt: Texture},
fonttitle=\small
]
\small\ttfamily
You are a professional image annotator. Please characterize the style attribute 'texture' of the input image a single phrase of four words (Do NOT use commas and word stacks), based on the following instructions:
\begin{enumerate}
    \item Identify the appearance and scale of repeating visual pattern (Describe the motif using only geometric or structural nouns); if no visible pattern is present, output "N/A".
    \item Describe the surface state, including the finish shaped by surface treatments, fabrication processes, and wear, as well as how the surface reflects and scatters light. Forbidden vague words: warm, soft, ornate, intricate, detailed, nice.
    \item Do not mention any content-related words or descriptions.
\end{enumerate}
\end{tcolorbox}
\noindent{\textbf{Medium.}}
Medium describes the materials or techniques used to create the reference image.
It encompasses physical media, such as watercolor, oil paint, charcoal, clay, plastic and embroidery, as well as computational techniques, such as pixel art and 3D rendering.
Beyond the basic medium, we further identify its fine-grained, medium-revealing visual cues to provide a more distinctive and specific characterization, such as ``long fur'' for ``3D rendering''.

\begin{tcolorbox}[
colback=gray!5,
colframe=gray!50,
boxrule=0.5pt,
arc=2pt,
left=6pt,
right=6pt,
top=6pt,
breakable,
bottom=6pt,
title=\textbf{Instruction Prompt: Medium},
fonttitle=\small
]
\small\ttfamily
You are a professional image annotator. Please identify the visible artistic medium of the input image based on the following instructions:
\begin{enumerate}
    \item Identify the materials or techniques used to create the input image. Avoid generic labels like "digital art/illustration".
    \item Further describe the medium's fine-grained visual cues. Avoid vibe words (such as soft, smooth, blended, realistic, intricate) or generic labels like "brushstrokes/lines", focusing on concrete, directly visible, medium-revealing evidence tokens (appearance of medium). For example, "long fur" and "pixel" (fine-grained visual cues) for "3D rendering" and "2D digital" (mediums).
    \item Output the fine-grained visual cues first, then the medium, in no more than four words total. Such as "long fur 3D rendering" and "pixel 2D digital".
    \item If no discriminative visual cues are visible, output the medium only. If the medium is not visually identifiable, output "N/A".
\end{enumerate}
\end{tcolorbox}
\noindent{\textbf{Brushwork.}}
Brushwork describes how visible stroke marks are applied and accumulated in an image, including their type (the technique or tool feel conveyed by visible brushstrokes, such as ``drybrush'', ``wash'', ``scrape'' and ``impasto''), scale (the apparent size/width/length of individual stroke marks) and layering (the buildup and layering depth of paint/marks).

\begin{tcolorbox}[
colback=gray!5,
colframe=gray!50,
boxrule=0.5pt,
arc=2pt,
left=6pt,
right=6pt,
top=6pt,
breakable,
bottom=6pt,
title=\textbf{Instruction Prompt: Brushwork},
fonttitle=\small
]
\small\ttfamily
You are a professional image annotator. Please characterize the style attribute 'brushwork' of the input image in a single phrase of four words (Do NOT use commas and word stacks) based on the following instructions:
\begin{enumerate}
    \item If no visible stroke marks are present, output ``N/A''.
    
    \item Otherwise, describe the brushwork in terms of type, scale, and layering according to the following definitions:
    \begin{itemize}
        \item \textbf{Type}: the technique or tool feel conveyed by visible brushstrokes, such as ``drybrush'', ``wash'', ``scrape'', and ``impasto'';
        \item \textbf{Scale}: the apparent size, width, or length of individual stroke marks;
        \item \textbf{Layering}: the buildup and layering depth of paint or marks.
    \end{itemize}
    
    \item Avoid vague descriptive words (e.g., ``soft'', ``intricate'', ``beautiful'', and ``smooth''); focus on concrete and directly observable properties.
    
    \item Output the scale and layering first, followed by the brushwork type, such as ``broad heavy impasto'' and ``fine flat wash''.
\end{enumerate}
\end{tcolorbox}

Finally, we identify whether visible outlines are present and, when present, describe their color and width. 
We also determine whether object edges are rendered as sharp boundaries or soft transitions.
We then combine these style annotations to construct style prompts according to the following template.
The components enclosed in square brackets are optional and are omitted when they are not visually identifiable.

\begin{tcolorbox}[
colback=gray!5,
colframe=gray!50,
boxrule=0.5pt,
arc=2pt,
left=6pt,
right=6pt,
top=6pt,
breakable,
bottom=6pt,
title=\textbf{Style Prompt Template},
fonttitle=\small
]
\small\ttfamily
[with <outline>]. In the style of <overall style identity>, <colors: dominant colors> with <colors: accent colors> accents, <lighting> lighting, [<texture: visual pattern> as repeating visual patterns,] <texture: surface state> surface, [<soft/sharp> edges,] [<brushwork>,] in <medium>.
\end{tcolorbox}

%% file: sec/4_experiments.tex
\begin{figure*}[t]
    \centering
    \includegraphics[width=0.9\linewidth]{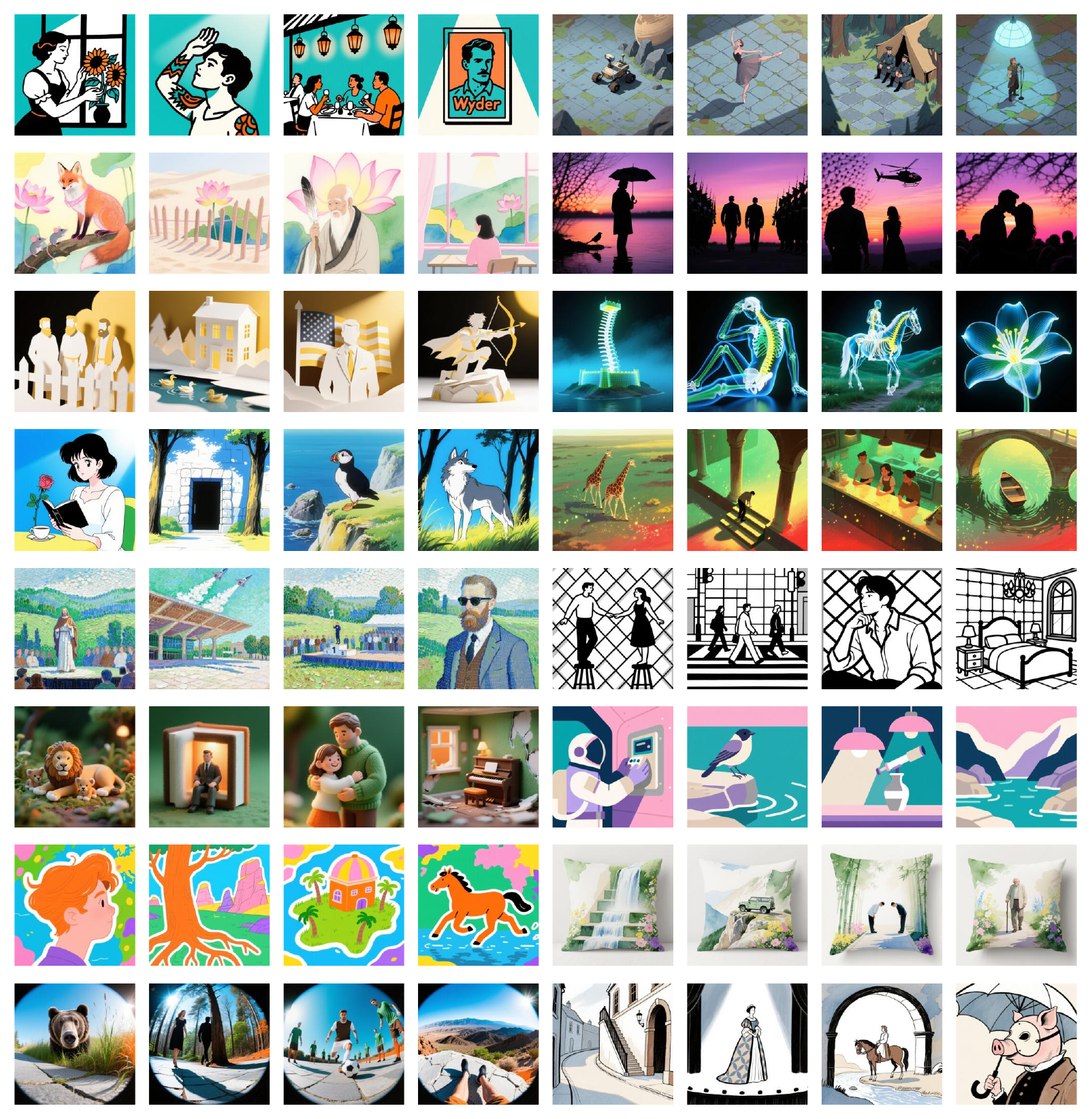}
    \vspace{-3mm}
    \caption{Visualizations of style pairs in MegaStyle++-8M. Each row presents two distinct style prompts, with four images generated from different semantic contents for each style, demonstrating strong intra-style consistency and diverse style coverage.}
    \vspace{-3mm}
    \label{fig:dataset}
\end{figure*}

\begin{table}[h]
    \centering
    \setlength{\tabcolsep}{1pt} 
    \setlength{\arrayrulewidth}{0.7pt}
    \caption{Comparison of style datasets. \cmark/\xmark \ indicate whether intra-style consistency is provided and \textemdash\ indicates that the statistic is unavailable.}
    \vspace{-2mm}
    \resizebox{\linewidth}{!}{
    \begin{tabular}{c|cccc}
        \toprule
       Datasets  & \makecell{Intra-style\\Consistency} & \makecell{Overall\\Style} & \makecell{Fine-grained\\Style} & \makecell{Style Image\\Number} \\
       \midrule
       WikiArt  & \xmark & 27 & \textemdash & 80K \\
       JourneyDB  & \xmark & \textemdash & 300K & 4.4M \\
       Style30K  & \xmark & \textemdash & 1K & 30K \\
       \midrule
       IMAGStyle  & \cmark & 14 & 15K & 210K \\
       OmniStyle-150K  & \cmark & \textemdash & 1K & 150K \\
       MegaStyle-1.4M  & \cmark & 8,355 & 170K & 1.4M \\
       \textbf{MegaStyle++-8M}  & \cmark & 150K & 1M & 8M \\
       \bottomrule
    \end{tabular}
    }
    \vspace{-2mm}
    \label{tab:data}
\end{table}

\section{Experiments}
\subsection{Implementation details}
We generate fine-grained and precise style annotations and construct MegaStyle++-8M following the data curation pipeline of MegaStyle \cite{gao2026megastyle}.
Specifically, based on the hierarchical style definition introduced in Section \ref{sec:method}, we employ Qwen3.5-35B-A3B\footnote{\url{https://huggingface.co/Qwen/Qwen3.5-35B-A3B}} to annotate 2M reference images from the Style Image Pool of MegaStyle using our refined instruction prompts for the overall style identity and style attributes.
Afterward, we assemble the structured style annotations into complete style prompts and apply ExactDeduplication, FuzzyDeduplication, and SemanticDeduplication from NeMo Curator to remove exact, near, and semantic duplicates from the resulting prompt gallery, retaining 1.5M prompts.
We then encode the deduplicated style prompts using mpnet and perform balanced sampling through a bottom-up three-level hierarchical (k)-means clustering with $k=\{100000, 5000, 500\}$, yielding 1M style prompts, which contain 150K unique overall style identity annotations.
Finally, we pair each style prompt with eight randomly sampled content prompts from the 400K content prompts in MegaStyle and use Qwen-Image to generate 8M stylized images, thereby constructing MegaStyle++-8M.
Table \ref{tab:data} compares MegaStyle++-8M with existing style datasets, including WikiArt \cite{phillips2011wiki}, JourneyDB \cite{sun2024journeydb}, Style30K \cite{li2024styletokenizer}, IMAGStyle \cite{xing2024csgo}, OmniStyle-150K \cite{wang2025omnistyle}, and MegaStyle \cite{gao2026megastyle}, in terms of intra-style consistency, numbers of overall style identities, numbers of fine-grained style, and dataset scale.
MegaStyle++-8M provides 150K overall style identities, 1M fine-grained and precise style prompts, and 8M stylized images, substantially exceeding all existing datasets in the compared dimensions and providing substantially broader coverage of the style space.
Visualizations of MegaStyle++-8M are presented in Figure \ref{fig:dataset}. 
Images generated from the same style prompt (each row) exhibit strong intra-style consistency across diverse semantic contents, while different style prompts capture a broad range of distinctive overall styles and fine-grained visual attributes.

\begin{figure}[t]
    \centering
    \includegraphics[width=0.9\linewidth]{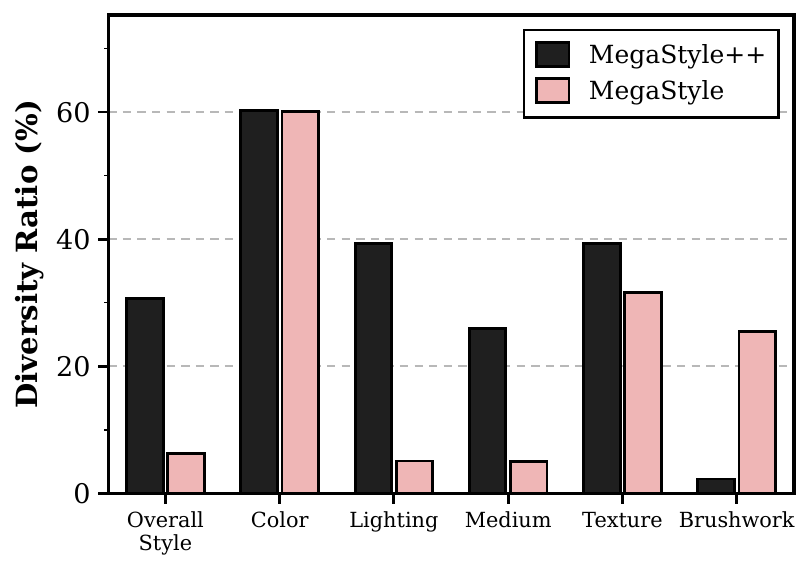}
    \vspace{-3mm}
    \caption{Diversity ratio of style annotations between MegaStyle++ and MegaStyle across different style attributes.}
    \vspace{-3mm}
    \label{fig:diverse}
\end{figure}

\begin{table}[t]
\centering
\caption{Top-3 most frequent annotations for lighting, medium, texture and brushwork.}
\vspace{-2mm}
\setlength{\tabcolsep}{4pt}
\renewcommand{\arraystretch}{1.12}

\resizebox{\linewidth}{!}{
\begin{tabular}{c|ll}
\toprule
\textbf{Attribute} & \textbf{MegaStyle++} & \textbf{MegaStyle} \\
\midrule

Lighting
&
\makecell[l]{
bright center fading to dark edges (1.08\%) \\
soft diffuse illumination from above (0.94\%) \\
left side bright fading right
 (0.82\%)
}
&
\makecell[l]{
diffused (9.47\%) \\
soft diffused (7.60\%) \\
even (7.40\%)
}
\\
\midrule

Medium
&
\makecell[l]{
photograph (2.00\%) \\
visible brushstrokes oil painting (1.53\%) \\
pixel 2d digital (1.15\%)
}
&
\makecell[l]{
digital rendering (26.63\%) \\
digital illustration (7.83\%) \\
digital painting (5.27\%)
}
\\
\midrule

Texture
&
\makecell[l]{
smooth matte finish (6.22\%) \\
flat matte finish (1.34\%) \\
smooth glossy finish (1.32\%)
}
&
\makecell[l]{
smooth texture (1.87\%) \\
smooth with subtle layering (1.34\%) \\
smooth reflective surfaces (1.09\%)
}
\\
\midrule

Brushwork
&
\makecell[l]{
fine flat wash (25.36\%) \\
fine layered drybrush (7.83\%) \\
fine layered wash (6.49\%)
}
&
\makecell[l]{
sharp edge hardness (1.69\%) \\
clean lines and sharp edges (1.39\%) \\
 clean and precise brushwork (1.30\%)
}
\\

\bottomrule
\end{tabular}
}

\label{tab:vibe}
\end{table}

\subsection{Further Analysis}
In this subsection, we compare our hierarchical style definition with the style formulation adopted in MegaStyle \cite{gao2026megastyle} to assess whether our definition produces more fine-grained and visually precise style descriptions while covering a more diverse style space.
We first quantify the diversity and semantic breadth of the style spaces induced by the two formulations through statistical and text-embedding-space analyses of the resulting style prompts.
Next, we examine whether the two formulations precisely capture the fine-grained style by comparing each reference image with the corresponding images generated from its style prompt.

\noindent\textbf{Diversity and Semantic Breadth.} 
We first randomly sample 100K reference images from the Style Image Pool and use the style instruction prompts of MegaStyle++ and MegaStyle to generate corresponding style annotations.
We then compute the proportion of unique annotations among the 100K samples as the diversity ratio ($\%$) for each style attribute, including overall style identity, color, lighting, medium, texture and brushwork.
As shown in Figure \ref{fig:diverse}, MegaStyle++ achieves substantially higher diversity ratios for overall style identity, lighting, and medium, while also improving texture diversity and maintaining comparable color diversity.

\begin{figure}[t]
    \centering
    \includegraphics[width=0.9\linewidth]{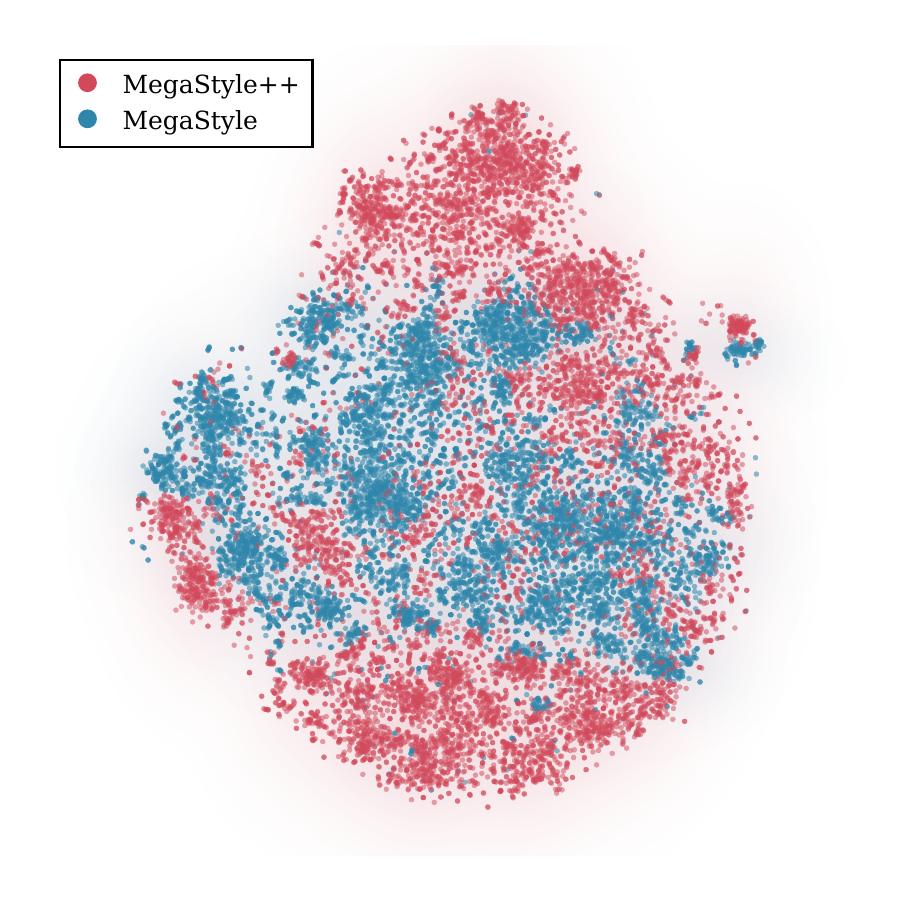}
    \vspace{-3mm}
    \caption{t-SNE \cite{van2008visualizing} visualization of SigLIP embeddings of 20K sampled style prompts from MegaStyle++ and MegaStyle. Each point represents one style prompt. MegaStyle++ exhibits a broader and more dispersed distribution, indicating greater diversity and semantic breadth.}
    \vspace{-3mm}
    \label{fig:fullprompts}
\end{figure}

\begin{figure*}
    \centering
    \includegraphics[width=0.9\linewidth]{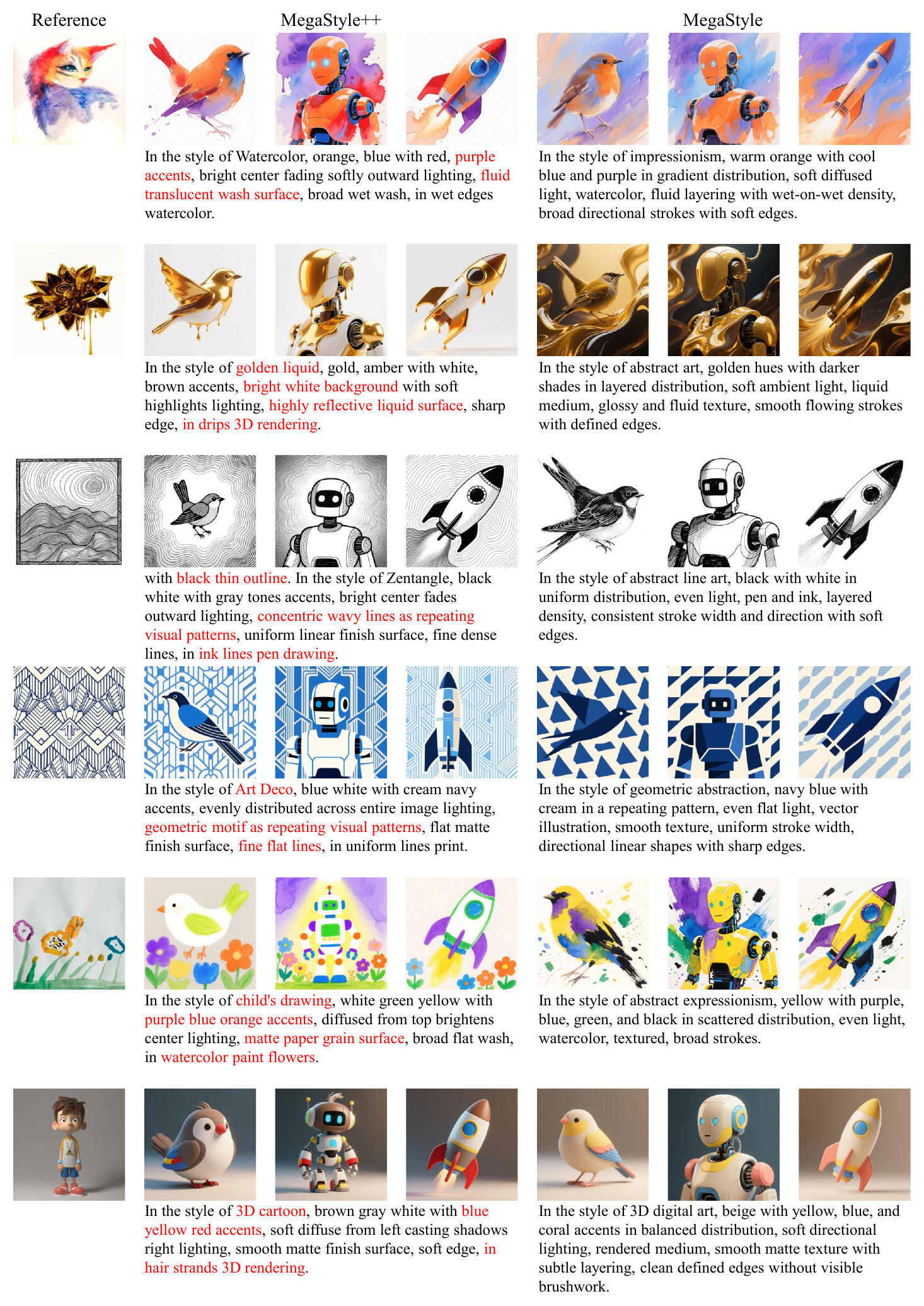}
    \vspace{-3mm}
    \caption{Qualitative comparison of style reproduction between MegaStyle++ and MegaStyle. The first column shows the real style reference images from StyleBench. The second and third columns present three reproduction images with the corresponding style prompts from MegaStyle++ and MegaStyle, respectively. From left to right, the three generated images correspond to the content prompts ``A bird'', ``A robot'', and ``A rocket''.}
    \vspace{-3mm}
    \label{fig:reproduces}
\end{figure*}

\begin{table*}[t]
\vspace{-2mm}
\caption{Quantitative comparison of style annotation quality and diversity between MegaStyle++ and MegaStyle. \textit{Relevance} and \textit{Vagueness} measure the attribute alignment and vague/vibe-based wording of style annotations using LLM. \textit{MPD}, \textit{Vendi}, and \textit{MNND} denote the average pairwise cosine distance, Vendi Score, and average nearest-neighbor distance in the SigLIP embedding space, respectively. MPD and MNND are reported after scaling by 100. Best results are marked in \textbf{bold}.}
\centering
\setlength{\arrayrulewidth}{0.7pt}
\setlength{\tabcolsep}{1pt}
\resizebox{\linewidth}{!}{
\begin{tabular}{c|cc|cc|cc|cc|cc|cc}
\toprule
&  \multicolumn{2}{c}{Overall Style}  & \multicolumn{2}{|c}{Color} & \multicolumn{2}{|c}{Lighting}  & \multicolumn{2}{|c}{Texture} & \multicolumn{2}{|c}{Medium}  & \multicolumn{2}{|c}{Brushwork}  \\
\midrule
{Metrics} &  MegaStyle++ & MegaStyle & MegaStyle++ & MegaStyle & MegaStyle++ & MegaStyle & MegaStyle++ & MegaStyle & MegaStyle++ & MegaStyle & MegaStyle++ & MegaStyle \\
\midrule
Relevance $\uparrow$  & \textbf{99.51} & 99.27 & \textbf{100.00} & 99.99  & \textbf{100.00} & 99.92 & \textbf{99.69} & 99.49 & \textbf{97.30} & 94.78  & \textbf{98.97} & 85.29\\
Vagueness $\downarrow$ & \textbf{17.10} & 24.50 & \textbf{0.05} & 9.18 & \textbf{12.00} & 78.20 & \textbf{16.90} & 39.20 & \textbf{1.40} & 3.50 & \textbf{21.10} & 37.10  \\
\midrule
MPD $\uparrow$ & \textbf{38.40} & 32.30 & 42.30 & \textbf{45.20} & \textbf{29.20} & 27.70 & \textbf{38.60} & 25.40 & \textbf{42.60} & 31.40 & \textbf{26.30} & 20.10 \\
Vendi $\uparrow$ & \textbf{13.60} & 7.33 & 7.92 & \textbf{10.85} & \textbf{6.53} & 4.88 & \textbf{11.92} & 5.26 & \textbf{14.52} & 5.16 & \textbf{4.17} & 3.64\\
MNND $\uparrow$ & \textbf{2.10} & 0.30 & 1.40 & \textbf{2.50} & \textbf{1.50} & 0.20 & \textbf{3.00} & 1.20 & \textbf{1.80} & 0.30 & 0.20 & \textbf{0.80}\\
\bottomrule
\end{tabular}
}
\vspace{-2mm}
\label{tab:main_result}
\end{table*}

\begin{table}[t]
    \centering
    \setlength{\tabcolsep}{2pt} 
    \setlength{\arrayrulewidth}{0.7pt}
    \caption{Quantitative comparison of style prompt diversity between MegaStyle++ and MegaStyle.}
    \vspace{-2mm}
    \resizebox{0.6\linewidth}{!}{
    \begin{tabular}{c|cc}
        \toprule
       Metric  & MegaStyle++ & MegaStyle \\
       \midrule
       MPD $\uparrow$  & \textbf{55.40} & 49.90  \\
       Vendi $\uparrow$  & \textbf{25.79} & 17.42\\
       MNND $\uparrow$ & \textbf{13.80} & 9.00  \\
       \bottomrule
    \end{tabular}
    }
    \vspace{-2mm}
    \label{tab:fullprompts}
\end{table}

Notably, we observe that MegaStyle tends to produce generic, vibe-based, or irrelevant annotations for lighting, medium, texture, and brushwork.
To provide a direct comparison, Table \ref{tab:vibe} lists the top-3 most frequent annotations for these style attributes generated by MegaStyle++ and MegaStyle.
For example, MegaStyle often relies on generic, vague, or impression-based wording, such as ``diffused'' and ``even'' for lighting, ``digital rendering'' for medium, and ``smooth texture'' for texture, rather than specifying concrete values of the aspects defined for each target attribute.
Moreover, although MegaStyle exhibits a higher diversity ratio for brushwork, its frequent annotations often describe other visual attributes, particularly edges and lines, rather than brushwork itself.
In contrast, MegaStyle++ produces more specific and aligned annotations by explicitly defining the style properties of each attribute.
We also employ an LLM as a judge to evaluate the relevance and vagueness of the generated style annotations on the full set of 100K reference images. 
For each style attribute, the judge is given its predefined definition and the corresponding description, and performs two binary evaluations: whether the description characterizes the defined attribute and whether it relies primarily on generic, subjective, or impression-based wording rather than concrete attribute values. 
The results are reported in the first two rows of Table \ref{tab:main_result}, where MegaStyle++ consistently produces more relevant and less vague descriptions across all style attributes.

To further evaluate diversity and semantic breadth, we encode the style annotations into the SigLIP~\cite{zhai2023sigmoid} embedding space and compute three complementary metrics: Mean Pairwise Distance (MPD) \cite{tang2025uncovering}, Vendi Score \cite{friedman2022vendi}, and Mean Nearest-Neighbor Distance (MNND) \cite{su2026emb}.
MPD is the average pairwise cosine distance between all annotation embeddings, measuring the overall spread of the annotations in the embedding space.
Vendi is computed as the exponential of the Shannon entropy \cite{shannon1948mathematical} of the eigenvalues of the normalized pairwise similarity matrix, measuring the effective diversity of the embedding distribution.
MNND is the average cosine distance from each annotation embedding to its nearest neighbor in the set, measuring the local separation between annotations.
As shown in the last three rows of Table~\ref{tab:main_result}, MegaStyle++ achieves substantially higher MPD, Vendi, and MNND scores for overall style identity, lighting, medium, and texture, while showing comparable performance for color, indicating broader semantic diversity across most style attributes.
Additionally, we compare the style prompts constructed by combining the annotations of all style attributes. The quantitative results in Table \ref{tab:fullprompts} and the visualization in Figure \ref{fig:fullprompts} demonstrate that MegaStyle++ provides more diverse style prompts than MegaStyle.

\noindent\textbf{Style Reproduction.}
To further evaluate the effectiveness of  our proposed hierarchical style definition in precisely capturing and reproducing the visual style of real reference images, we compare it with its counterpart in MegaStyle through a style reproduction experiment.
Specifically, we caption real style reference images from StyleBench \cite{11165480} using the two formulations to construct their corresponding style prompts.
Each style prompt is then combined with the same set of content prompts, including ``A bench'', ``A bird'', ``A rocket'', ``A car'', and ``A robot'', and fed into Qwen-Image \cite{wu2025qwen} to generate the corresponding reproduction images.
As shown in Figure \ref{fig:reproduces}, the reproduction results of MegaStyle++ precisely capture both the overall style identity and fine-grained style attributes.
For example, in row 2, it successfully identifies the ``golden liquid'' style with a white background and dripping 3D-rendered appearance, which are consistent with the reference image.
It also captures the distinctive wavy-line and geometric-motif repeating textures in rows 3 and 4.
In contrast, MegaStyle struggles to capture these intrinsic visual cues and often produces generic or inaccurate style descriptions.
We further measure the style similarity between the reference and reproduced images using CSD \cite{somepalli2024measuring} and MegaStyle-Encoder \cite{gao2026megastyle}, as well as the text-image similarity between the style prompts and reference images using CLIP \cite{radford2021learning}.
As shown in Table~\ref{tab:reproduces}, MegaStyle++ consistently outperforms MegaStyle across all three metrics, demonstrating that our hierarchical style definition provides more accurate style descriptions and enables more faithful style reproduction.

\begin{table}[h]
    \centering
    \setlength{\tabcolsep}{2pt} 
    \setlength{\arrayrulewidth}{0.7pt}
    \caption{Quantitative comparison of style reproduction between MegaStyle++ and MegaStyle.}
    \vspace{-2mm}
    \resizebox{0.75\linewidth}{!}{
    \begin{tabular}{c|cc}
        \toprule
       Metric  & MegaStyle++ & MegaStyle \\
       \midrule
       CSD $\uparrow$  & \textbf{50.19} & 33.54  \\
       MegaStyle-Encoder $\uparrow$  & \textbf{58.40} & 40.92\\
       \midrule
       CLIP-Text $\uparrow$ & \textbf{24.49} & 20.71  \\
       \bottomrule
    \end{tabular}
    }
    \vspace{-2mm}
    \label{tab:reproduces}
\end{table}

%% file: sec/5_conclusion.tex
\section{Conclusion}
In this work, we revisit the fundamental question of how image style should be defined and propose a unified hierarchical style definition that organizes style from an overall style identity to fine-grained visual attributes.
Based on this definition, we construct MegaStyle++-8M with 150K overall style identities, 1M fine-grained style prompts, and 8M stylized images, substantially expanding the breadth and granularity of the image style space.
Extensive experiments demonstrate that MegaStyle++ provides more diverse, precise, and definition-aligned style descriptions than MegaStyle, while enabling more faithful reproduction of real reference styles. 
Beyond dataset scaling, our hierarchical definition provides an explicit and interpretable language for describing transferable visual style, offering a practical foundation for future research in style representation, understanding, and generation.

%% file: main.bib
@String(AAAI = {AAAI})

@inproceedings{everaert2023diffusion,
  title={Diffusion in style},
  author={Everaert, Martin Nicolas and Bocchio, Marco and Arpa, Sami and S{\"u}sstrunk, Sabine and Achanta, Radhakrishna},
  booktitle={Proceedings of the IEEE/CVF International Conference on Computer Vision},
  pages={2251--2261},
  year={2023}
}

@inproceedings{lu2023specialist,
  title={Specialist Diffusion: Plug-and-Play Sample-Efficient Fine-Tuning of Text-to-Image Diffusion Models To Learn Any Unseen Style},
  author={Lu, Haoming and Tunanyan, Hazarapet and Wang, Kai and Navasardyan, Shant and Wang, Zhangyang and Shi, Humphrey},
  booktitle={Proceedings of the IEEE/CVF Conference on Computer Vision and Pattern Recognition},
  pages={14267--14276},
  year={2023}
}

@inproceedings{gatys2016image,
  title={Image style transfer using convolutional neural networks},
  author={Gatys, Leon A and Ecker, Alexander S and Bethge, Matthias},
  booktitle={Proceedings of the IEEE conference on computer vision and pattern recognition},
  pages={2414--2423},
  year={2016}
}

@article{ulyanov2016texture,
  title={Texture networks: Feed-forward synthesis of textures and stylized images},
  author={Ulyanov, Dmitry and Lebedev, Vadim and Vedaldi, Andrea and Lempitsky, Victor},
  journal={arXiv preprint arXiv:1603.03417},
  year={2016}
}

@inproceedings{johnson2016perceptual,
  title={Perceptual losses for real-time style transfer and super-resolution},
  author={Johnson, Justin and Alahi, Alexandre and Fei-Fei, Li},
  booktitle={Computer Vision--ECCV 2016: 14th European Conference, Amsterdam, The Netherlands, October 11-14, 2016, Proceedings, Part II 14},
  pages={694--711},
  year={2016},
  organization={Springer}
}

@article{dumoulin2016learned,
  title={A learned representation for artistic style},
  author={Dumoulin, Vincent and Shlens, Jonathon and Kudlur, Manjunath},
  journal={arXiv preprint arXiv:1610.07629},
  year={2016}
}

@inproceedings{huang2017arbitrary,
  title={Arbitrary style transfer in real-time with adaptive instance normalization},
  author={Huang, Xun and Belongie, Serge},
  booktitle={Proceedings of the IEEE international conference on computer vision},
  pages={1501--1510},
  year={2017}
}

@article{sohn2024styledrop,
  title={Styledrop: Text-to-image synthesis of any style},
  author={Sohn, Kihyuk and Jiang, Lu and Barber, Jarred and Lee, Kimin and Ruiz, Nataniel and Krishnan, Dilip and Chang, Huiwen and Li, Yuanzhen and Essa, Irfan and Rubinstein, Michael and others},
  journal={Advances in Neural Information Processing Systems},
  volume={36},
  year={2024}
}

@inproceedings{ruiz2023dreambooth,
  title={Dreambooth: Fine tuning text-to-image diffusion models for subject-driven generation},
  author={Ruiz, Nataniel and Li, Yuanzhen and Jampani, Varun and Pritch, Yael and Rubinstein, Michael and Aberman, Kfir},
  booktitle={Proceedings of the IEEE/CVF Conference on Computer Vision and Pattern Recognition},
  pages={22500--22510},
  year={2023}
}

@article{gal2022image,
  title={An image is worth one word: Personalizing text-to-image generation using textual inversion},
  author={Gal, Rinon and Alaluf, Yuval and Atzmon, Yuval and Patashnik, Or and Bermano, Amit H and Chechik, Gal and Cohen-Or, Daniel},
  journal={arXiv preprint arXiv:2208.01618},
  year={2022}
}

@article{li2017universal,
  title={Universal style transfer via feature transforms},
  author={Li, Yijun and Fang, Chen and Yang, Jimei and Wang, Zhaowen and Lu, Xin and Yang, Ming-Hsuan},
  journal={Advances in neural information processing systems},
  volume={30},
  year={2017}
}

@article{wang2023styleadapter,
  title={Styleadapter: A single-pass lora-free model for stylized image generation},
  author={Wang, Zhouxia and Wang, Xintao and Xie, Liangbin and Qi, Zhongang and Shan, Ying and Wang, Wenping and Luo, Ping},
  journal={arXiv preprint arXiv:2309.01770},
  year={2023}
}

@article{liu2023stylecrafter,
  title={StyleCrafter: Enhancing Stylized Text-to-Video Generation with Style Adapter},
  author={Liu, Gongye and Xia, Menghan and Zhang, Yong and Chen, Haoxin and Xing, Jinbo and Wang, Xintao and Yang, Yujiu and Shan, Ying},
  journal={arXiv preprint arXiv:2312.00330},
  year={2023}
}

@inproceedings{radford2021learning,
  title={Learning transferable visual models from natural language supervision},
  author={Radford, Alec and Kim, Jong Wook and Hallacy, Chris and Ramesh, Aditya and Goh, Gabriel and Agarwal, Sandhini and Sastry, Girish and Askell, Amanda and Mishkin, Pamela and Clark, Jack and others},
  booktitle={International conference on machine learning},
  pages={8748--8763},
  year={2021},
  organization={PMLR}
}

@article{ye2023ip,
  title={Ip-adapter: Text compatible image prompt adapter for text-to-image diffusion models},
  author={Ye, Hu and Zhang, Jun and Liu, Sibo and Han, Xiao and Yang, Wei},
  journal={arXiv preprint arXiv:2308.06721},
  year={2023}
}

@article{sun2024journeydb,
  title={Journeydb: A benchmark for generative image understanding},
  author={Sun, Keqiang and Pan, Junting and Ge, Yuying and Li, Hao and Duan, Haodong and Wu, Xiaoshi and Zhang, Renrui and Zhou, Aojun and Qin, Zipeng and Wang, Yi and others},
  journal={Advances in Neural Information Processing Systems},
  volume={36},
  year={2024}
}

@article{phillips2011wiki,
  title={Wiki Art Gallery, Inc.: A case for critical thinking},
  author={Phillips, Fred and Mackintosh, Brandy},
  journal={Issues in Accounting Education},
  volume={26},
  number={3},
  pages={593--608},
  year={2011},
  publisher={American Accounting Assocation}
}

@inproceedings{zhang2023inversion,
  title={Inversion-based style transfer with diffusion models},
  author={Zhang, Yuxin and Huang, Nisha and Tang, Fan and Huang, Haibin and Ma, Chongyang and Dong, Weiming and Xu, Changsheng},
  booktitle={Proceedings of the IEEE/CVF Conference on Computer Vision and Pattern Recognition},
  pages={10146--10156},
  year={2023}
}

@article{qi2024deadiff,
  title={DEADiff: An Efficient Stylization Diffusion Model with Disentangled Representations},
  author={Qi, Tianhao and Fang, Shancheng and Wu, Yanze and Xie, Hongtao and Liu, Jiawei and Chen, Lang and He, Qian and Zhang, Yongdong},
  journal={arXiv preprint arXiv:2403.06951},
  year={2024}
}

@inproceedings{yang2023zero,
  title={Zero-shot contrastive loss for text-guided diffusion image style transfer},
  author={Yang, Serin and Hwang, Hyunmin and Ye, Jong Chul},
  booktitle={Proceedings of the IEEE/CVF International Conference on Computer Vision},
  pages={22873--22882},
  year={2023}
}

@inproceedings{ahn2024dreamstyler,
  title={Dreamstyler: Paint by style inversion with text-to-image diffusion models},
  author={Ahn, Namhyuk and Lee, Junsoo and Lee, Chunggi and Kim, Kunhee and Kim, Daesik and Nam, Seung-Hun and Hong, Kibeom},
  booktitle={Proceedings of the AAAI Conference on Artificial Intelligence},
  volume={38},
  number={2},
  pages={674--681},
  year={2024}
}

@article{sun2024create,
  title={Create your world: Lifelong text-to-image diffusion},
  author={Sun, Gan and Liang, Wenqi and Dong, Jiahua and Li, Jun and Ding, Zhengming and Cong, Yang},
  journal={IEEE Transactions on Pattern Analysis and Machine Intelligence},
  year={2024},
  publisher={IEEE}
}

@article{croitoru2023diffusion,
  title={Diffusion models in vision: A survey},
  author={Croitoru, Florinel-Alin and Hondru, Vlad and Ionescu, Radu Tudor and Shah, Mubarak},
  journal={IEEE Transactions on Pattern Analysis and Machine Intelligence},
  volume={45},
  number={9},
  pages={10850--10869},
  year={2023},
  publisher={IEEE}
}

@article{xia2024diffusion,
  title={A diffusion model translator for efficient image-to-image translation},
  author={Xia, Mengfei and Zhou, Yu and Yi, Ran and Liu, Yong-Jin and Wang, Wenping},
  journal={IEEE Transactions on Pattern Analysis and Machine Intelligence},
  year={2024},
  publisher={IEEE}
}

@article{xia2024diffi2i,
  title={Diffi2i: efficient diffusion model for image-to-image translation},
  author={Xia, Bin and Zhang, Yulun and Wang, Shiyin and Wang, Yitong and Wu, Xinglong and Tian, Yapeng and Yang, Wenming and Timotfe, Radu and Van Gool, Luc},
  journal={IEEE Transactions on Pattern Analysis and Machine Intelligence},
  year={2024},
  publisher={IEEE}
}

@article{hu2021lora,
  title={Lora: Low-rank adaptation of large language models},
  author={Hu, Edward J and Shen, Yelong and Wallis, Phillip and Allen-Zhu, Zeyuan and Li, Yuanzhi and Wang, Shean and Wang, Lu and Chen, Weizhu},
  journal={arXiv preprint arXiv:2106.09685},
  year={2021}
}

@inproceedings{zhai2023sigmoid,
  title={Sigmoid loss for language image pre-training},
  author={Zhai, Xiaohua and Mustafa, Basil and Kolesnikov, Alexander and Beyer, Lucas},
  booktitle={Proceedings of the IEEE/CVF international conference on computer vision},
  pages={11975--11986},
  year={2023}
}

@ARTICLE{11165480,
  author={Gao, Junyao and Sun, Yanan and Liu, Yanchen and Tang, Yinhao and Zeng, Yanhong and Qi, Ding and Chen, Kai and Zhao, Cairong},
  journal={IEEE Transactions on Pattern Analysis and Machine Intelligence}, 
  title={StyleShot: a Snapshot on any Style}, 
  year={2025},
  volume={},
  number={},
  pages={1-15},
  doi={10.1109/TPAMI.2025.3610614}}

@inproceedings{wang2025omnistyle,
  title={OmniStyle: Filtering High Quality Style Transfer Data at Scale},
  author={Wang, Ye and Liu, Ruiqi and Lin, Jiang and Liu, Fei and Yi, Zili and Wang, Yilin and Ma, Rui},
  booktitle={Proceedings of the Computer Vision and Pattern Recognition Conference},
  pages={7847--7856},
  year={2025}
}

@article{wu2025qwen,
  title={Qwen-image technical report},
  author={Wu, Chenfei and Li, Jiahao and Zhou, Jingren and Lin, Junyang and Gao, Kaiyuan and Yan, Kun and Yin, Sheng-ming and Bai, Shuai and Xu, Xiao and Chen, Yilei and others},
  journal={arXiv preprint arXiv:2508.02324},
  year={2025}
}

@article{xing2024csgo,
  title={Csgo: Content-style composition in text-to-image generation},
  author={Xing, Peng and Wang, Haofan and Sun, Yanpeng and Wang, Qixun and Bai, Xu and Ai, Hao and Huang, Renyuan and Li, Zechao},
  journal={arXiv preprint arXiv:2408.16766},
  year={2024}
}

@inproceedings{li2024styletokenizer,
  title={Styletokenizer: Defining image style by a single instance for controlling diffusion models},
  author={Li, Wen and Fang, Muyuan and Zou, Cheng and Gong, Biao and Zheng, Ruobing and Wang, Meng and Chen, Jingdong and Yang, Ming},
  booktitle={European Conference on Computer Vision},
  pages={110--126},
  year={2024},
  organization={Springer}
}

@article{somepalli2024measuring,
  title={Measuring style similarity in diffusion models},
  author={Somepalli, Gowthami and Gupta, Anubhav and Gupta, Kamal and Palta, Shramay and Goldblum, Micah and Geiping, Jonas and Shrivastava, Abhinav and Goldstein, Tom},
  journal={arXiv preprint arXiv:2404.01292},
  year={2024}
}

@article{gao2026megastyle,
  title={Megastyle: Constructing diverse and scalable style dataset via consistent text-to-image style mapping},
  author={Gao, Junyao and Liu, Sibo and Li, Jiaxing and Sun, Yanan and Tu, Yuanpeng and Shen, Fei and Zhang, Weidong and Zhao, Cairong and Zhang, Jun},
  journal={arXiv preprint arXiv:2604.08364},
  year={2026}
}

@article{wu2024fiva,
  title={Fiva: Fine-grained visual attribute dataset for text-to-image diffusion models},
  author={Wu, Tong and Xu, Yinghao and Po, Ryan and Zhang, Mengchen and Yang, Guandao and Wang, Jiaqi and Liu, Ziwei and Lin, Dahua and Wetzstein, Gordon},
  journal={Advances in Neural Information Processing Systems},
  volume={37},
  pages={31990--32011},
  year={2024}
}

@inproceedings{ruta2022stylebabel,
  title={Stylebabel: Artistic style tagging and captioning},
  author={Ruta, Dan and Gilbert, Andrew and Aggarwal, Pranav and Marri, Naveen and Kale, Ajinkya and Briggs, Jo and Speed, Chris and Jin, Hailin and Faieta, Baldo and Filipkowski, Alex and others},
  booktitle={European Conference on Computer Vision},
  pages={219--236},
  year={2022},
  organization={Springer}
}

@inproceedings{tang2025uncovering,
  title={Uncovering the bigger picture: Comprehensive event understanding via diverse news retrieval},
  author={Tang, Yixuan and Shi, Yuanyuan and Sun, Yiqun and Tung, Anthony Kum Hoe},
  booktitle={Proceedings of the 2025 Conference on Empirical Methods in Natural Language Processing},
  pages={33927--33945},
  year={2025}
}

@article{friedman2022vendi,
  title={The vendi score: A diversity evaluation metric for machine learning},
  author={Friedman, Dan and Dieng, Adji Bousso},
  journal={arXiv preprint arXiv:2210.02410},
  year={2022}
}

@article{su2026emb,
  title={emb-diversity: A Tool for Embedding-Based Measurement of Data Diversity},
  author={Su, Cantao and Velayuthan, Menan and Ploeger, Esther and Nguyen, Dong and Wegmann, Anna},
  journal={arXiv preprint arXiv:2607.19848},
  year={2026}
}

@article{shannon1948mathematical,
  title={A mathematical theory of communication},
  author={Shannon, Claude Elwood},
  journal={The Bell system technical journal},
  volume={27},
  number={3},
  pages={379--423},
  year={1948},
  publisher={Nokia Bell Labs}
}

@article{van2008visualizing,
  title={Visualizing data using t-SNE.},
  author={Van der Maaten, Laurens and Hinton, Geoffrey},
  journal={Journal of machine learning research},
  volume={9},
  number={11},
  year={2008}
}
